\documentclass[10pt, a4paper]{article}
\usepackage{authblk}
\usepackage{bm}
\usepackage{enumitem}

\usepackage[]{lrec-coling2024} 
\usepackage{graphicx}
\usepackage{algorithm}
\usepackage{algorithmic}
\usepackage{times}
\usepackage{latexsym}
\usepackage{graphicx}
\usepackage{pgfplots}
\usepackage{amsmath}
\usepackage{bm}
\usepackage{url}
\usepackage{latexsym}
\usepackage{multirow}
\usepackage{colortbl}
\usepackage{booktabs}
\usepackage{amssymb}
\usepackage{amsmath}
\usepackage{amsthm}
\usepackage{graphicx}
\usepackage{makecell}
\usepackage{threeparttable}
\usepackage{subcaption}
\usepackage{fancyhdr}
\usepackage{float}
\usepackage{supertabular}
\usepackage{tablefootnote}
\usepackage{multirow}
\usepackage{microtype}
\usepackage{amsmath, amssymb, amscd, amsfonts}
\usepackage{IEEEtrantools}
\usepackage{soul}
\usepackage{tikz}
\usepackage{pgfplots}
\usepackage{verbatim}
\usepackage{url}
\usepackage{algorithm}
\usepackage{algorithmic}
\usepackage{amssymb}
\usepackage{amsfonts} 
\usepackage{multirow}
\usepackage{boldline}
\usepackage{hhline}
\usepackage{natbib} 
\usepackage{amssymb}
\usepackage{amsfonts} 
\usepackage{bm}
\usepackage{multirow}
\usepackage{boldline}
\usepackage{hhline}
\usepackage{amsmath}
\usepackage{pifont}
\usepackage{makecell}
\usepackage{mathrsfs}
\usepackage{utfsym}
\usepackage[normalem]{ulem}
\useunder{\uline}{\ul}{}
\usepackage{latexsym}
\usepackage{newfloat}
\usepackage{listings}

\title{Let’s Rectify Step by Step: Improving Aspect-based Sentiment Analysis with Diffusion Models}

\name{\normalsize Shunyu Liu$^1$, Jie Zhou$^{1,*}$\thanks{$^*$ Corresponding author, jzhou@cs.ecnu.edu.cn.}, Qunxi Zhu$^{2}$, Qin Chen$^1$, Qingchun Bai$^3$, Jun Xiao$^3$, Liang He$^1$} 

\address{\normalsize $^1$ School of Computer Science and Technology, East China Normal University, Shanghai, China \\
         $^2$ Research Institute of Intelligent Complex Systems, Fudan University, Shanghai, China \\
         $^3$ Shanghai Engineering Research Center of Open Distance Education, Shanghai Open University \\
}

\abstract{
Aspect-Based Sentiment Analysis (ABSA) stands as a crucial task in predicting the sentiment polarity associated with identified aspects within text. However, a notable challenge in ABSA lies in precisely determining the aspects' boundaries (start and end indices), especially for long ones, due to users' colloquial expressions. 
We propose \texttt{DiffusionABSA}, a novel diffusion model tailored for ABSA, which extracts the aspects progressively step by step. Particularly, \texttt{DiffusionABSA} gradually adds noise to the aspect terms in the training process, subsequently learning a denoising process that progressively restores these terms in a reverse manner. To estimate the boundaries, we design a denoising neural network enhanced by a syntax-aware temporal attention mechanism to chronologically capture the interplay between aspects and surrounding text. Empirical evaluations conducted on eight benchmark datasets underscore the compelling advantages offered by \texttt{DiffusionABSA} when compared against robust baseline models. Our code is publicly available at \url{https://github.com/Qlb6x/DiffusionABSA}.
 \\ \newline \Keywords{Diffusion Models, Aspect-based Sentiment Analysis, Syntax} }

\begin{document}

\maketitleabstract

\section{Introduction}
Aspect-Based Sentiment Analysis (ABSA) \cite{fan2018multi}, a prominent text analysis technique of the past decade, has garnered significant research attention. ABSA involves extracting aspect terms and discerning the sentiment associated with each aspect. 
The ABSA landscape encompasses four pivotal sentiment subtasks, namely aspect term extraction (AE), aspect category detection, opinion term extraction (OE), and sentiment classification (SC) \cite{zhang2022survey}. Within this paper, our primary focus centers on two of these subtasks: AE and SC, collectively referred to as AESC.
For example, in the sentence ``{Amazing \textit{Spanish Mackeral special appetizer} and perfect \textit{box sushi} ( that eel with avodcao -- um um um ).}'', there are two sets of aspect terms and their sentiment polarities, (\textit{Spanish Mackeral special appetizer}, Positive) and  (\textit{box sushi}, Positive). 


Some previous studies perform AE and SC independently in a pipeline, potentially leading to error propagation \cite{fan2019target,hu2019open}. Recently, end-to-end models are designed to address two subtasks jointly via unified tagging schema \cite{mitchell2013open,zhang2015neural,li2019unified} or machine reading comprehension framework \cite{yang2022aspect}. 
Furthermore, generative techniques have emerged as a powerful tool in tackling ABSA challenges \cite{zhang2022survey,yan2021unified}.
These methods often involve the generation of sentiment element sequences adhering to specific formats, thereby harnessing the nuances of label semantics.
However, a recurring limitation across these approaches lies in their struggle to precisely delineate the boundaries of aspects due to the inherent diversity of language expression. 
The ambiguity and fluidity of language usage can lead to indistinct boundaries, where one aspect term might encompass multiple words (e.g., ``Spanish Mackeral special appetizer''), and a single sentence could encompass multiple aspect terms. 


\begin{table*}[h!]
\centering
\scriptsize
\begin{tabular}{lc}
\hlineB{3}
 STEP 1  & Amazing $\overbrace{\mathrm{\colorbox{green!30}{Spanish Mackeral special appetizer}}}^{\mathrm{ASPECT~\textcolor{green}{\usym{2713}}} \mathrm{POSITIVE~\textcolor{green}{\usym{2713}}} }$ and perfect$\overbrace{\mathrm{\colorbox{orange!30}{box sushi}}}^{\mathrm{MISSING~\textcolor{red}{\usym{2717}}}}$ ( that eel with avodcao -- um um um ).   \\\hline
 STEP 2  & Amazing $\overbrace{\mathrm{\colorbox{red!30}{Spanish Mackeral special appetizer and perfect box sushi}}}^{\mathrm{ASPECT~\textcolor{red}{\usym{2717}}} \mathrm{POSITIVE~\textcolor{green}{\usym{2713}}}}$ ( that eel with avodcao -- um um um ).   \\\hline
 STEP 3  & Amazing $\overbrace{\mathrm{\colorbox{green!30}{Spanish Mackeral special appetizer}}}^{\mathrm{ASPECT~\textcolor{green}{\usym{2713}}}\mathrm{POSITIVE~\textcolor{green}{\usym{2713}}}}$ and perfect$\overbrace{\mathrm{\colorbox{green!30}{box sushi}}}^{\mathrm{ASPECT~\textcolor{green}{\usym{2713}}}\mathrm{POSITIVE~\textcolor{green}{\usym{2713}}}}$ ( that eel with avodcao -- um um um ).  \\ \hline
  & ... \\ \hline
  \hlineB{3}
\end{tabular}
\caption{The boundary of aspect terms gradually changes during the denoising process in \texttt{DiffusionABSA}. The spans annotated with \colorbox{green!30}{green}, \colorbox{orange!30}{orange}, and \colorbox{red!30}{red} respectively signify the correct, missing, wrong results.}
\label{table:introducefig2}
\end{table*}

In response to this intricate challenge, we introduce an innovative solution by integrating a diffusion model \cite{sohl2015deep} – a paradigm that has showcased impressive capabilities in controlled generation tasks. Notably, diffusion models have demonstrated remarkable performance in various domains, including text-to-image generation \cite{zhang2023text, nichol2022glide}, and text generation \cite{nachmani2021zero, he2022diffusionbert}. These achievements stand as testaments to the potential of diffusion models in facilitating token-level controls \cite{zou2023diffusion}.
At its core, a diffusion model orchestrates the process of generation in a stepwise manner. During training, it introduces noise to the input, progressively refining the generation process. Subsequently, it learns a complementary denoising procedure to accurately restore the original input. Leveraging this underlying mechanism, we propose an ingenious fusion of the diffusion model with ABSA, harnessing its capabilities to enhance the inference of the aspects. 

This paper introduces \texttt{DiffusionABSA}, a novel architecture for diffusion architecture for ABSA which effectively marries the controlled generation process of diffusion models with the intricate aspect detection challenge characteristic of ABSA. 
\texttt{DiffusionABSA} is structured around two fundamental processes: corruption and denoising. 
The corruption process gradually adds Gaussian noise to the aspect terms according to a fixed variance schedule. 
The denoising process undoes the added noise at each time step iteratively and learns to faithfully reconstruct the original data by reversing this noising process. 
Illustrative insights from Table \ref{table:introducefig2} underscore the distinctive features of \texttt{DiffusionABSA} in handling intricate aspect boundaries. Specifically, the model demonstrates iterative refinement in aspect extraction: initially missing an aspect (``box sushi''), then combining two aspects into one (``Spanish Mackeral special appetizer and perfect box sushi''), and finally, accurately extracting both aspects (``Spanish Mackeral special appetizer'' and ``box sushi'').
To further bolster \texttt{DiffusionABSA}'s capabilities, we introduce a denoising neural network equipped with a syntax-aware temporal attention strategy. 
This strategic augmentation facilitates the model's adeptness in capturing the temporal evolution of aspect-text interactions, resulting in a more effective aspect modeling process.

To comprehensively assess the efficacy of our proposed \texttt{DiffusionABSA}, we conduct a series of experiments across eight diverse datasets, benchmarked against several state-of-the-art (SOTA) baselines. The empirical findings affirm the superiority of \texttt{DiffusionABSA} in most cases. Additionally, ablation studies provide nuanced insights into the contributions of key components within our model, further validating its effectiveness in addressing the ABSA challenge.

The principal contributions are summarized as follows.
\begin{itemize}[leftmargin=*, align=left]
    \item We propose \texttt{DiffusionABSA}, a novel framework that adapts diffusion models to refine the aspect progressively through a dynamic interplay of corruption and denoising processes.
    \item We design a denoising neural network enhanced by a syntax-aware temporal attention strategy, which estimates the boundaries temporally in the reverse diffusion process. 
    \item A series of experiments on eight widely-used benchmark datasets show that \texttt{DiffusionABSA} achieves new SOTA performance in most cases. Notably, our model showcases superior performance over ChatGPT, highlighting its efficacy in ABSA.
\end{itemize}

\begin{figure*}[!t]
\centering
\includegraphics[width=0.95\textwidth]{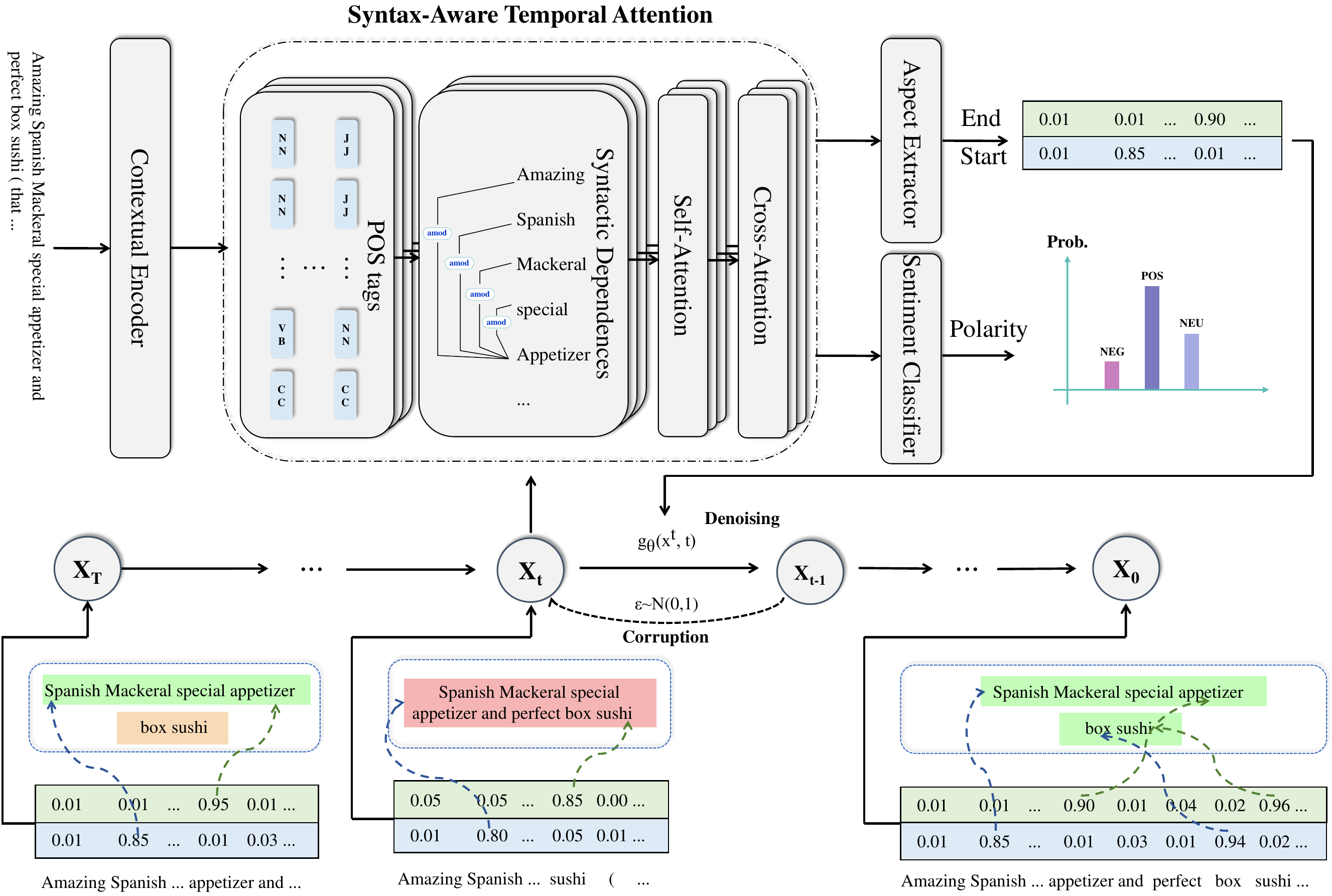} 
\caption{The framework of \texttt{DiffusionABSA}.}
\label{fig: framework}
\end{figure*}

\section{Related Work}
\subsection{Aspect-based Sentiment Analysis}
Aspect-Based Sentiment Analysis (ABSA) is a pivotal endeavor that identifies sentiment-related components within a sentence \cite{schouten2015survey,ma2019exploring,li2020conditional,zhou2020sk}. These components encompass aspects, opinions, and sentiments, collectively contributing to a comprehensive understanding of the textual content.
In the nascent stages of ABSA research, the focus predominantly gravitated toward individual subtasks, namely aspect terms extraction (AE), opinion extraction (OE), or sentiment classification (SC) \cite{zhang2022survey,zhou2020sentix}. 
This includes the convergence of multiple subtasks, such as the simultaneous treatment of aspect extraction and sentiment classification (AESC) \cite{yan2021unified}, aspect-opinion pair extraction (AOPE) \cite{fan2019target} and triplet extraction (TE) \cite{peng2020knowing}, to model the relationships among them. 

This study primarily centers its attention on AESC. Previous efforts, exemplified by \newcite{hu2019open, zhou2019span}, introduced span-based AESC techniques that amalgamated AESC at the span level. 
These models that merged AE and SC in a pipeline framework are susceptible to error propagation, as highlighted by \newcite{hu2019open}. 
Recent efforts \cite{li2019exploiting, li2019unified} have aimed to handle the entire ABSA task using end-to-end models and a unified tagging schema.
Nonetheless, the intricacies of language expression's diversity often hinder the accurate detection of aspects in these models.
This paper contributes a novel \texttt{DiffusionABSA} framework that tactfully integrates diffusion models to advance ABSA by modeling the aspects progressively.

\subsection{Diffusion Models}
Diffusion models \cite{sohl2015deep,ho2020denoising,song2020denoising}  which have recently emerged as a new one of SOTA generative models, have achieved impressive synthesis results on image data. 
Denoising diffusion probabilistic models (DDPMs) were initially introduced by \newcite{sohl2015deep}, and \newcite{ho2020denoising} brought theoretical breakthroughs and innovations. DDPMs contain two major processes: adding noise in a forward diffusion process and restoring the original data in a denoising process. 
The forward process corrupts information by gradually adding Gaussian noise: 
\begin{equation}
q(\bm{x}_{1:T}|\bm{x}_{0})=\prod_{t=1}^{T}q(\bm{x}_{t}|\bm{x}_{t-1})  \label{eq}
\end{equation}
\begin{equation}
q(\bm{x}_{t}|\bm{x}_{t-1})=\mathcal{N}(\bm{x}_{t};\sqrt{1-\beta_{t}}\bm{x}_{t-1},\beta_{t}\bm{I}).  \label{eq}
\end{equation}
\par
Notably, the forward process allows sampling $\bm{x}_{t}$ at an arbitrary time step $t$ directly based on the initial data sample $\bm{x}_0$:
\begin{equation}
q(\bm{x}_{t}|\bm{x}_{0})=\mathcal{N}(\bm{x}_{t};\sqrt{\bar{\alpha}_{t}}\bm{x}_{0},(1-\bar{\alpha}_{t})\bm{I})   \label{eq}
\end{equation}
where $\alpha_{t}=1-\beta_{t}$ and $\bar{\alpha}_{t}=\prod_{t=1}^{T}\alpha_{t}$.

Diffusion models learn the reverse process to restore the original data step by step. If $\beta _{t}$ is small enough, that is, the added noise at each step is relatively small the reverse process can be modeled as a Markov chain with learned conditional Gaussians \cite{song2020denoising}, parameterized by a neural network:
\begin{equation}
p_{\bm{\theta}}(\bm{x}_{0:T}) = p(\bm{x}_T) \prod^T_{t=1} p_{\bm{\theta}}(\bm{x}_{t-1} \vert \bm{x}_t) \label{eq}
\end{equation}
\begin{equation}
\\p_{\bm{\theta}}(\bm{x}_{t-1} \vert \bm{x}_t) = \mathcal{N}(\bm{x}_{t-1}; \boldsymbol{\mu}_{\bm{\theta}}(\bm{x}_t, t), \boldsymbol{\Sigma}_{\bm{\theta}}(\bm{x}_t, t))   \label{eq}
\end{equation}
where $\bm{\mu}_{\bm{\theta}}(\bm{x}_t, t)$ and $\boldsymbol{\Sigma}_{\bm{\theta}}(\bm{x}_t, t)$ is the predicted covariance and mean of $p_{\bm{\theta}}(\bm{x}_{t-1} \vert \bm{x}_t)$ computed by a neural network.

Recently, the application of the diffusion model in the realm of natural language processing (NLP) has garnered notable attention from researchers, as evident in works such as \newcite{nachmani2021zero, he2022diffusionbert}. This application can be broadly categorized into two directions: 1) Continuous Diffusion Models: This line of research involves encoding discrete tokens into a continuous space and subsequently executing both the forward and reverse diffusion processes \cite{li2022diffusion}; 2) Discrete Diffusion Models: Operating within the discrete token space, this direction extends the principles of diffusion models to encompass discrete state-spaces \cite{reid2022diffuser}.
Leveraging the potent capabilities of the diffusion model in a controlled generation, our paper introduces \texttt{DiffusionABSA} to tackle the intricate challenges of aspect extraction and sentiment classification.

\section{Our Proposed Framework}
In this paper, we propose \texttt{DiffusionABSA}, which learns a denoising neural network for ABSA based on a diffusion 
architecture (Figure \ref{fig: framework}). 
First, \texttt{DiffusionABSA} adds noise $\bm{\epsilon}$ into aspect representation at each step in the forward corruption process.
Next, \texttt{DiffusionABSA} recovers the input using a denoising neural network $g_{\bm{\theta}}$ in the backward denoising process. 
Particularly, we design a denoising neural network with a syntax-aware temporal attention mechanism to estimate the boundaries $g_{\bm{\theta}}(\bm{x}_t, t, {S})$ at the time step $t$.

Formally, ABSA aims to extract aspects and the corresponding sentiment polarities $Y = \{(a_i,p_{i})\}_{i=1}^{|Y|}$ from the given sentence ${S}=\{w_{1},w_{2},...w_{|S|}\}$, where $a_i$ and $p_{i}$ are the $i$-th aspect and its sentiment polarity (neutral, positive or negative). $|Y|$ is the number of aspects in sentence $S$ and $|S|$ is the length of the sentence.  Each aspect $a_{i}=(s_{i},e_{i})$ is defined by all the tokens between $s_{i}$ and $e_{i}$, where $s_{i}$ and $e_{i}$ are the start and end indices of aspect $a_i$, $1\leq s_{i}\leq e_{i}\leq |S|$.

\subsection{Aspect Diffusion}
As shown in Table \ref{table:introducefig2}, we extract the aspects progressively via a diffusion model under the Markov chain assumption, where each step $\bm{x}_t$ only depends on the previous step $\bm{x}_{t-1}$ with $T$ steps. \texttt{DiffusionABSA} adds noise to the aspects to model $p(\bm{x}_t|\bm{x}_{t-1})$ in the forward corruption process while recovering them from the noise to model $q_{\bm{\theta}}(\bm{x}_{t-1}|\bm{x}_t)$ in the backward denoising process. 

\paragraph{Forward Corruption Process}
The forward corruption process is a process of adding a small amount of Gaussian noise by a fixed schedule to the aspect term boundaries step by step. 
We normalize the start and end indices as initial step $\bm{x}_0$:
\begin{equation}
    \bm{x}_0=\lambda\left(\frac{[[s_1, e_1], ..., [s_i, e_i], [s_{N}, e_{N}]]}{|S|}-0.5\right)  
\end{equation}
where $\lambda$ is a hyper-parameter to scale the value to $(-\lambda, \lambda)$ and $N$ is the max $|Y|$ in the dataset. 

Instead of applying the forward process multiple times repeatedly to get the desired data point $\bm{x}_{t}$ at $t<T$, we apply a simple reparameterization to get the desired output by precomputing the variances and certain parameters:
\begin{eqnarray}
\bm{x}_{t}& = &\sqrt{\alpha_{t}}\bm{x}_{t-1}+\sqrt{1-\alpha_{t}} \bm{\epsilon}_{t-1}\nonumber\\
& = &\sqrt{\alpha_{t}\alpha_{t-1}}\bm{x}_{t-2} + \sqrt{1-\alpha_{t}\alpha_{t-1}} \bar{\bm{\epsilon}}_{t-2}\nonumber\\
& = &\text{...}\nonumber\\
& = &\sqrt{\bar{\alpha}_{t}}\bm{x}_{0}+\sqrt{1-\bar{\alpha}_{t}}\bm{\epsilon}\label{eq}
\end{eqnarray}
where $\alpha_{t}=1-\beta_{t}$, $\bar{\alpha}_{t}=\prod_{t=1}^{T}\alpha_{t}$ and $\bm{\epsilon} \sim \mathcal{N}(\bm{0}, \bm{I})$.
In accordance with this equation, we can directly compute the $\bm{x}_t$ from $\bm{x}_0$ in the corruption process.

\paragraph{Backward Denoising Process}
During the backward process, \texttt{DiffusionABSA} 
progressively refines the aspect boundaries by the learned denoising process. Specifically, we restore the $\bm{x}_{0}$ from the noisy $\bm{x}_{T}$ based on the conditional Gaussian $p_{\bm{\theta}}(\bm{x}_{t-1}|\bm{x}_t)$:
\begin{equation}
\small
\begin{aligned}
  \bm{x}_{t-1} = &  \sqrt{\alpha_{t-1}}\cdot g_{\bm{\theta}}(\bm{x}_t, t, {S}) \\
& + \sqrt { 1 - \alpha _ { t - 1 }}\cdot \frac{\bm{x}_t - \sqrt{\alpha_{t}}g_{\bm{\theta}}(\bm{x}_t, t, {S})}{\sqrt{1-\alpha _ { t}}} + \sigma _ { t } \cdot \bm{\epsilon}_{t}  
\end{aligned}
\end{equation}
where $g_{\bm{\theta}}(\bm{x}_{t}, t, S)$ represents a denoising neural network used to predict the distribution of start and end indices, where $\bm{\theta}$ is the learnable parameters of the denoising neural network. 

To expedite the reverse process of \texttt{DiffusionABSA}, we adopt DDIMs \cite{song2020denoising} which is a straightforward scheduler that converts the stochastic process into a deterministic one, requiring a small number of sampling steps. 
DDIMs construct a class of non-Markovian diffusion processes that lead to the same training objective, but whose reverse process can be much faster to sample from. We define $\bm{\tau}$ as an increasing sub-sequence of $[1,2,…,T]$, where $\gamma$ is the length of $\bm{\tau}$ and $\bm{\tau}_{\gamma}=T$:
\begin{equation}
\scriptsize
\nonumber
\begin{aligned}
  \bm{x}_{\bm{\tau}_{t-1}} = &  \sqrt{\alpha_{\bm{\tau}_{t-1}}} g_{\bm{\theta}}(\bm{x}_{\bm{\tau}_t}, \bm{\tau}_t, {S}) \\
& + \sqrt { 1 - \alpha _ { \bm{\tau}_{t - 1} }}
\frac{\bm{x}_{\bm{\tau}_t} - \sqrt{\alpha_{\bm{\tau}_t}}g_{\bm{\theta}}(\bm{x}_{\tau_t}, \bm{\tau}_t, {S})}{\sqrt{1-\alpha _ {\bm{\tau}_{t }}}} + \sigma _ { \bm{\tau}_t } \bm{\epsilon}_{\bm{\tau}_t}  
\end{aligned}
\end{equation}
where $\sigma_{\bm{\tau}_t}$ is commonly chosen as zero.
Finally, to calculate the boundaries of aspect terms at each step, we design a denoising neural network $g_{\bm{\theta}}(\bm{x}_{t}, t, {S})$.  

\subsection{Denoising Neural Network}
This section presents a denoising neural network $g_{\bm{\theta}}(\bm{x}_{t},t,{S})$, comprising four components: contextual encoder, syntax-aware temporal attention, aspect extractor, and sentiment classifier.
Utilizing the sentence representation learned by the contextual encoder, we introduce a syntax-aware temporal attention mechanism to sequentially model connections between aspects and text.
Then, we use an aspect extractor and sentiment classifier to predict the boundaries and sentiments of the aspects based on the sentence representation learned by syntax-aware temporal attention.

\paragraph{Contextual Encoder}
Pre-trained language models (PLM) \cite{devlin2018bert,liu2019roberta} have been shown prominent in retrieving the contextualized features for various NLP tasks. Hence we utilize PLM (e.g., BERT, RoBERTa) as the underlying encoder to encode the sentence $S$ into vectors:
\begin{equation}
\boldsymbol{H} = \mathrm{Encoder}(\{w_{1},w_{2},...,w_{|S|}\})\label{eq}.
\end{equation}
\par
The start of sentence $([\mathrm{CLS}])$ and the end of sentence $([\mathrm{SEP}])$ tokens are added to the start and end of $S$, which are disregarded in the equations for the sake of simplification.

\paragraph{Syntax-aware Temporal Attention (SynTA)}
Within the denoising process, we devise a syntax-aware temporal attention method tailored for the temporal inference of aspect boundaries.
This involves combining part-of-speech (POS) tags and dependency trees to capture the interaction between text and aspects. Furthermore, we incorporate time features to capture the temporal information.

To initiate, we get POS tags $\{p_{1},p_{2},...,p_{|S|}\}$ by StandfordCoreNLP tool, where $p_{i}$ is word $w_{i}$' POS tag. POS features hold significant importance in identifying potential boundaries, thereby guiding the model to appropriately identify aspects. We use a learnable POS embedding layer to embed the POS tags, where $\bm{E}^p \in\mathbb{R}^{|S|\times d}$: 
\begin{equation}
\boldsymbol{E}^p = \mathrm{POSEmbedding}(\{p_{1},p_{2},...,p_{|S|}\})\label{eq}.
\end{equation}

To further integrate the syntactic dependency, we adopt a graph convolution network (GCN) model. 
We define an undirected graph $G=<V, E>$ with self-loop edges, where $E$ is a list of dependency edges between each pair of words and $V$ is a list of words. The adjacency matrix $\mathbf{M}\in\mathbb{R}^{|S|\times |S|}$ is defined as follows:
\begin{equation}
\boldsymbol{M}_{ij}=\begin{cases}l_{i,j},&\text{if a dependency edge between }w_i ,w_j,\\0,&\text{otherwise},\end{cases}\label{eq}
\end{equation}
where $l_{i,j}$ is the index of dependency label between $w_i$ and $w_j$.
Then, the hidden representation of the word $w_{i}$ at the $k$-th layer of GCNs is computed as:
\begin{equation}
\nonumber
\small
\boldsymbol{H}_i^{k}=\mathrm{ReLU}\left(\sum_{j=1}^{|S|}u^{k-1}_{i, j}\cdotp(\boldsymbol{W_1}\cdotp[\boldsymbol{H}_j^{k-1}; \boldsymbol{E}^d_{i,j};\boldsymbol{E}^p_j]+b)\right)
\end{equation}
\begin{equation}
u_{i, j}^{k-1}=\boldsymbol{M}_{i,j}\cdotp\mathrm{Softmax}(\boldsymbol{W_2}\cdotp[\boldsymbol{H}_j^{k-1}; \boldsymbol{E}^d_{i,j};\boldsymbol{E}^p_j])
\end{equation}
where $\boldsymbol{E}^d_{i,j}$ is the embedding of dependency label $l_{i,j}$, $\boldsymbol{H}_i^{0}$ is the word embedding $H_i$ learned by contextual encoder. $\boldsymbol{W}_1$ and $\boldsymbol{W}_2$ are learnable parameters.
The final representation is defined as $\hat{\boldsymbol{H}}=\boldsymbol{H}^{K}$, where $K$ is the number of layers.

Based on syntax-enhanced representation $\hat{\boldsymbol{H}}$, we obtain aspect representation $\boldsymbol{H}^a$ via average pooling.
In order to explore the internal connection between sentence representations $\hat{\boldsymbol{H}}$ and span representations $\boldsymbol{H}^a$, 
we utilize a self-attention and a cross-attention layer to model the interaction.
\begin{equation}
\hat{\boldsymbol{H}}^a = \mathrm{SelfAttention}(\boldsymbol{W}_1^Q \boldsymbol{H}^a,\boldsymbol{W}_1^K \boldsymbol{H}^a,\boldsymbol{W}_1^V \boldsymbol{H}^a)
\end{equation}
\begin{equation}
\hat{\boldsymbol{H}}^a=\mathrm{CrossAttention}(\boldsymbol{W}_2^Q \hat{\boldsymbol{H}}^a, \boldsymbol{W}_2^K\hat{\boldsymbol{H}}, \boldsymbol{W}_2^V\hat{\boldsymbol{H}}).
\end{equation}

To consider the timestep, we incorporate the sinusoidal embedding $\boldsymbol{E}_t$ of timestep $t$ using sine and cosine functions \cite{DBLP:conf/nips/VaswaniSPUJGKP17}. The final time-related representations of the aspects are calculated as follows:
\begin{equation}
\hat{\boldsymbol{H}}^a=\mathrm{TimeEmbedding}(\hat{\boldsymbol{H}}^a, \boldsymbol{E}_t)
\end{equation}
where $\mathrm{TimeEmbedding}(\cdotp,\cdotp)$ is an operation to combine two vector representations. In this paper, we use addition or multiple interactions with scale and shift \cite{dumoulin2018feature, perez2018film}; other operations can also be used.

\paragraph{Aspect Extractor}
We use an aspect extractor to predict the start and end indices of the aspects.
 In precise terms, we calculate the probability of the aspect span boundary $\bm{P}^{b}$ and $b\in \{\mathrm{start}, \mathrm{end}\}$ with $\hat{\boldsymbol{H}}^a$ and $\hat{\boldsymbol{H}}$:
\begin{equation}
\bm{P}^{b}=\mathrm{Sigmoid}(\hat{\boldsymbol{H}}^a \boldsymbol{W^b_a} + \hat{\boldsymbol{H}} \boldsymbol{W^b_s}). \label{eq}
\end{equation}

\paragraph{Sentiment Classifier}
And probability $\bm{P}^{y}$  of the sentiment polarity, $y \in (\mathrm{positive}, \mathrm{negative}, \mathrm{neutral})$ is calculated by the classifier:
\begin{equation}
\bm{P}^{y}=\mathrm{Softmax}(\mathrm{MLP}(\hat{\boldsymbol{H}}^a)). \label{eq}
\end{equation}

Utilizing the aforementioned extractor and classifier, we can decode the predicted probabilities of $\bm{P}^{\mathrm{start}}$, $\bm{P}^{\mathrm{end}}$ and $\bm{P}^{y}$ to get the outputs $Y_i=((s_i, e_i), p_i)$ for the $i$-th aspect. 

\paragraph{Loss Function}
DiffusionABSA progressively refines the aspect boundaries based on $\bm{x_t}$ through the learned denoising process. Like traditional DDPMs, loss will be computed in each intermediate step. We use the cross-entropy (CE) losses between the ground truth $((s_i, e_i), p_i)$ and predicted aspect probabilities of the left and right boundary indexes and type of entity, $P_i^{start}, P_i^{end}, P_i^{y}$ :
\begin{equation}
\small
\mathcal{L}=-\sum_{i=1}^K (CE(P_i^{start}, s_i) + CE(P_i^{end}, e_i) + CE(P_i^{y}, p_i))
\end{equation}

\section{Experiment Setting}
\subsection{Datasets and Metrics}
We conducted an extensive evaluation using four distinct ABSA datasets, each comprising two versions denoted as $D_{20a}$ \cite{peng2020knowing}, $D_{20b}$ \cite{xu2020position}, which contain restaurant (res) and laptop (lap) reviews. 
Detailed statistical summaries of these datasets can be found in Table \ref{tab:statistics}. 
Following \newcite{yan2021unified}, we employ micro-F1 scores as the evaluation metric in our experiments. 
In our evaluation methodology, the correctness of a predicted aspect term and its associated sentiment polarity is contingent upon the precise alignment of its span with the corresponding boundaries delineated by the ground truth. Additionally, the polarity classification must concord with the actual sentiment polarity. 
This rigorous evaluation approach ensures a stringent validation of the predictive efficacy of our proposed method.

\begin{table}[]
\centering
\small
\setlength{\tabcolsep}{0.8mm}{\begin{tabular}{lrrrrrrrr}
\hlineB{4}
\multirow{2}{*}{\textbf{Dataset}}     & \multicolumn{2}{c}{\textbf{14res}}          & \multicolumn{2}{c}{\textbf{14lap}}          & \multicolumn{2}{c}{\textbf{15res}}          & \multicolumn{2}{c}{\textbf{16res}}          \\ \cline{2-9} 
                      & \textbf{\#S}   & \textbf{\#T}      & \textbf{\#S}   & \textbf{\#T}      & \textbf{\#S}   & \textbf{\#T}      & \textbf{\#S}   & \textbf{\#T}      \\ \hline
\multicolumn{9}{c}{\textbf{$D_{20a}$}}
\\ \hline
Train                 & 1300                 & 2145                 & 920                  & 1265                 & 593                  & 923                  & 842                  & 1289                 \\
Dev                   & 323                  & 524                  & 228                  & 337                  & 148                  & 238                  & 210                  & 316                  \\
Test                  & 496                  & 862                  & 339                  & 490                  & 318                  & 455                  & 320                  & 465                  \\ \hline
\multicolumn{9}{c}{\textbf{$D_{20b}$}}
\\ \hline
Train                 & 1266                 & 2338                 & 906                  & 1460                 & 605                  & 1013                 & 857                  & 1394                 \\
Dev                   & 310                  & 577                  & 219                  & 346                  & 148                  & 249                  & 210                  & 339                  \\
Test                  & 492                  & 994                  & 328                  & 543                  & 322                  & 485                  & 326                  & 514                  \\ 
\hlineB{4}
\end{tabular}}
\caption{Statistics of the datasets pertinent to the ABSA task. \#S and \#T denote the quantities of sentences and targets within the respective datasets.}
\label{tab:statistics}
\end{table}

\subsection{Baselines}
To ensure a comprehensive comparative analysis, we have meticulously outlined the most proficient baseline models for AE and AESC subtasks. This comprehensive delineation is aimed at facilitating a thorough assessment of the proposed experimental framework against existing benchmarks, including the pipeline, joint, end-to-end, and large language models. 

We first compare our \texttt{DiffusionABSA} against the pipeline methods: 
\begin{itemize}[leftmargin=*, align=left]
    \item \textbf{SPAN-BERT}~\cite{hu2019open} is a pipeline method for AESC which takes BERT as the backbone network. A span boundary detection model is used for AE subtask, and then followed by a polarity classifier for SC.
    \item \textbf{Peng-two-stage}~\cite{peng2020knowing} is a two-stage pipeline model. Peng-two-stage extracts both aspect-sentiment pairs and opinion terms in the first stage. In the second stage, a classiﬁer is used to ﬁnd the valid pairs from the first stage and ﬁnally construct the triplet prediction.
\end{itemize}

Some methods investigate ABSA using joint models:
\begin{itemize}[leftmargin=*, align=left]
    \item \textbf{MIN} \cite{yu2021making} is a multi-task learning method named to make flexible use of subtasks for a unified ABSA. 
    \item \textbf{SPAN} \cite{hu2019open} is a span-based extract-then-classify model, where opinions are directly extracted from the sentence based on supervised target boundaries and corresponding polarities are then classified by extracted spans. 
    \item \textbf{Dual-MRC}~\cite{mao2021joint} is a joint training model that incorporates two machine reading comprehension (MRC) modules used separately for AE and AESC.
\end{itemize}

The summation of significant findings from end-to-end ABSA endeavors is detailed below:

\begin{itemize}[leftmargin=*, align=left]
    \item \textbf{SynGen}~\cite{yu2023syngen} adds syntactic inductive bias to attention assignment and thus directs attention to the correct target words which apply to AESC, Pair, and Triplet subtasks.
    \item \textbf{SyMux}~\cite{fei2022inheriting} is a multiplex decoding method. It improves the framework by using syntactic information to identify term boundaries and pairings, transferring sentiment layouts and clues from simpler tasks to more challenging ones.
    \item \textbf{Li-unified}~\cite{li2019unified} addresses target-based sentiment analysis as a complete task in an end-to-end manner. The model introduces a novel unified tagging scheme to achieve this goal.
    \item \textbf{RINANTE+} \cite{peng2020knowing}, is modified from the work \cite{ma2018joint}. RINANTE+ is an LSTM-CRF model which first uses dependency relations of words to extract opinions and aspects with the sentiment. Then, all the candidate aspect-opinion pairs with position embedding are fed into the Bi-LSTM encoder to make a final classification.
    \item \textbf{CMLA+} \cite{peng2020knowing} is adjusted from the one \cite{wang2017coupled} which is an attention-based model following the same two-stage processing with dependency relations as RINANTE+.
    \item \textbf{GEN} \cite{yan2021unified} converts all ABSA subtasks into a unified generative formulation, and provides a real unified end-to-end solution for the whole ABSA subtasks, which could benefit multiple tasks. 
    \item \textbf{GTS} \cite{wu2020grid} is a tagging scheme different from pipeline methods, to address the Aspect-oriented Fine-grained Opinion Extraction (AFOE) task in an end-to-end fashion only with one unified grid tagging task.
    \item \textbf{RCAL} \cite{chen2020relation} allows the subtasks to work coordinately via the multi-task learning and relation propagation mechanisms in a stacked multi-layer network.
\end{itemize}

Furthermore, our scope of comparative analysis was broadened to encompass Large Language Models (LLMs):

\begin{itemize}[leftmargin=*, align=left]
    \item \textbf{ChatGPT} \cite{openai2023} is one of the best-known examples of LLMs from OpenAI’s GPT (Generative Pre-Training Transformer) series, and is capable of generating human-like text based on context and past conversations.
    Additionally, we explore zero-shot (Zero-shot) prompts, few-shot in-context learning (ICL) prompts~\cite{brown2020language}, and chain-of-thought (COT) prompts~\cite{wei2022chain} settings.
\end{itemize}

\begin{table}[t!]
\centering
\small
\setlength{\tabcolsep}{1.2mm}{
\begin{tabular}{lcccc}
\hlineB{4}
\textbf{MODEL}                   & \textbf{14res} & \textbf{14lap} & \textbf{15res} & \textbf{16res} \\ \hline
CMLA+                            & 70.62          & 56.90          & 53.60          & 61.20          \\ 
RINANTE+                         & 48.15          & 36.70          & 41.30          & 42.10          \\ 
Li-unified                       & 73.79          & 63.38          & 64.95          & 70.20          \\ 
Peng-two-stage                   & 74.19          & 62.34          & 65.79          & 71.73          \\ 
Dual-MRC                         & 76.57          & 64.59          & 65.14          & 70.84          \\ 
SPAN-BART                        & 78.47          & 68.17          & 69.95          & 75.69          \\ 
SyMux                            & 78.68          & 70.32          & 69.08          & 77.95          \\ 
SynGen                           & {79.72 }         & 70.06          & 71.61          & 77.51          \\ 
ChatGPT (Zero-shot)                       &  59.08          & 45.48          & 53.91          & 55.40          \\
ChatGPT (5-shot ICL)                       & 65.98         & 49.50         & 63.66          & 63.11          \\ 
ChatGPT (5-shot COT)                       & 62.82          & 48.87          & 66.07          & 65.93          \\ \hline
DiffusionABSA                    &\textbf{80.93} & \textbf{72.81} & \textbf{76.70} & \textbf{81.72} \\ 
\multicolumn{1}{l}{~~~~~w/o SynTA} &80.84         &72.39            &74.26            &80.53                \\ \hlineB{4}
\end{tabular}}
\caption{Results of AESC over $D_{20a}$ datasets. We use the results of baselines reported in \newcite{yu2023syngen}.}
\label{tab:d20a-1}
\end{table}

\subsection{Implementation}
Our experimental evaluations encompass eight benchmark datasets. For the AE and AESC tasks, we incorporate dependency trees and POS tags as an integral preprocessing step. We adopt the RoBERTa as our pre-trained language model. The optimization process is orchestrated using the AdamW optimizer, initialized with a learning rate of 0.0002. All experiments are conducted on a potent 24GB RTX3090 GPU. The training regimen, comprising 100 epochs, is concluded within an hour, employing a batch size of 16. This streamlined experimental setup enables us to efficiently explore the model's performance across various tasks and datasets.

\section{Results and Analyses} 

\subsection{Main Results}
In Tables \ref{tab:d20a-1} and \ref{tab:d20a-2}, we present the results of \texttt{DiffusionABSA} on dataset $D_{20a}$ as well as the baselines on AE and AESC tasks. Based on the outcomes, we deduce the ensuing observations. 
\textbf{First}, our method achieves significant improvements against almost all the baselines on micro-F1 score. 
Impressively, our approach outperforms the SOTA model, exhibiting an average improvement of 3.31\% on AESC. We also extend our analysis to the D$_{20b}$ dataset, as illustrated in Table \ref{tab:d20b}. Similar trends emerge, as our methodology consistently demonstrates a competitive edge over baselines. This consistent pattern of superior performance substantiates our model's capacity to leverage the SynTA encoder effectively, resulting in the proficient distinction of aspect term representations. 
\textbf{Second}, it is worth noting that in comparison to the performance reported in~\cite{han2023information} for ChatGPT, which incorporates ICL prompts and COT prompts, our \texttt{DiffusionABSA} stands out as a robust contender. Despite the reported enhancements achieved by ChatGPT through these prompt-based strategies, it remains evident that ChatGPT falls short of both SOTA and \texttt{DiffusionABSA} in terms of competitive performance and consistent achievement.

\begin{table}[t!]
\centering
\small
\setlength{\tabcolsep}{1.2mm}{
\begin{tabular}{lcccc}
\hlineB{4}
\textbf{MODEL}                   & \textbf{14res} & \textbf{14lap} & \textbf{15res}         & \textbf{16res}        \\ \hline
SPAN                             & 86.71          & 82.34          & 74.63                  & 74.68                 \\ 
RACL                             & 86.38          & 81.79          & 73.99                  & 74.91                 \\ 
MIN                              & 87.91          & 83.22          & - & - \\ 
CMLA                             & 81.22          & 79.53          & 76.03                  & 74.20                 \\ 
RINANTE                          & 81.34          & 80.40          & 73.38                  & 72.82                 \\ 
Li-unified                       & 81.62          & 78.56          & 74.65                  & 73.36                 \\ 
GTS                              & 83.82          & 82.48          & 78.22                  & 75.80                 \\  
GEN                              & 87.07          & 83.52          & 75.48                  & 81.35                 \\ 
SyMux                            & \textbf{89.02}          & 84.42          & 79.73                  & 82.41                 \\ 
ChatGPT (Zero-shot)                       &  55.65          & 43.03          & 40.33          & -          \\
ChatGPT (5-shot ICL)                       & 70.99         & 48.19         & 53.49          & -          \\ 
ChatGPT (5-shot COT)                         & 72.41          & 54.50          & 59.27                  & - \\  \hline
DiffusionABSA                    & {87.15} & \textbf{86.66} & \textbf{85.40}         & \textbf{87.87}        \\
\multicolumn{1}{l}{~~~~~w/o SynTA} &86.93   &86.01         &83.15      &86.23                       \\ \hlineB{4}
\end{tabular}}
\caption{Results of AE over $D_{20a}$ datasets. We use the results of baselines reported in \newcite{fei2022inheriting}.} 
\label{tab:d20a-2}
\end{table}

\begin{table*}[]
\centering
\small
\setlength{\tabcolsep}{1.0mm}{
\begin{tabular}{lcccccccc}
\hlineB{4}
\multicolumn{1}{c}{\multirow{2}{*}{\textbf{MODEL}}} & \multicolumn{4}{c}{\textbf{14res}}                                                                                                & \multicolumn{4}{c}{\textbf{16res}}                                                                                                \\ \cline{2-9} 
\multicolumn{1}{c}{}                                & \multicolumn{1}{l}{ALL} & \multicolumn{1}{l}{LEN=1} & \multicolumn{1}{l}{LEN=2} & \multicolumn{1}{l}{LEN\textgreater{}2} & \multicolumn{1}{l}{ALL} & \multicolumn{1}{l}{LEN=1} & \multicolumn{1}{l}{LEN=2} & \multicolumn{1}{l}{LEN\textgreater{}2} \\ \hline
SeqLab                                 & 66.17                   & 58.88                     & 16.11                     & 4.36                                   & 68.60                   & 58.94                     & 19.37                     & 4.73                                   \\
DiffusionABSA                          & 79.13                   & 71.68                     & 20.82                     & 7.51                                   & 78.87                   & 68.97                     & 21.61                     & 8.49                                   \\ \hline
Improvement           & 19.59\%                  & 21.74\%                    & 29.24\%                     & 72.25\%                                  & 14.97\%                   & 17.02\%                   & 11.56\%                     & 79.49\%                                  \\ \hlineB{4}
\end{tabular}}
\caption{Results over aspects with various lengths on $D_{20a}$ datasets.}
\label{tab:length}
\end{table*}

\subsection{Ablation Studies}
As delineated in Tables \ref{tab:d20a-1}, \ref{tab:d20a-2}, and \ref{tab:d20b}, we conduct the ablation studies by removing the SynTA (w/o SynTA) on the AE and AESC tasks. Upon meticulous experimentation, we observed a compelling trend: the direct omission of the SynTA module has a discernible adverse impact on model performance, reflected in a reduction of the F1 score over all the datasets. On datasets D$_{20a}$ and D$_{20b}$, this omission respectively led to an average decline of 1.04\% and 1.76\%. These empirically substantiated findings reaffirm the pivotal role played by the SynTA strategy in fortifying the interaction between aspect terms and sentences. The synthesis of these outcomes underscores the effectiveness of SynTA in augmenting the cohesive integration of aspect terms within sentences, thus contributing significantly to the overall model performance.

\begin{table}[]
\centering
\scriptsize
\setlength{\tabcolsep}{1.2mm}{
\begin{tabular}{lllllll}
\hlineB{4}
\multicolumn{1}{c}{\multirow{2}{*}{\textbf{MODEL}}} & \multicolumn{2}{c}{\textbf{14res}}                                  & \multicolumn{2}{c}{\textbf{14lap}}                                  & \multicolumn{2}{c}{\textbf{15res}}               \\ \cline{2-7} 
\multicolumn{1}{c}{}                                 & \multicolumn{1}{c}{\textbf{AE}} & \multicolumn{1}{c}{\textbf{AESC}} & \multicolumn{1}{c}{\textbf{AE}} & \multicolumn{1}{c}{\textbf{AESC}} & \multicolumn{1}{c}{\textbf{AE}} & \textbf{AESC}  \\ \hline
SPAN-BERT                                            & 86.71                           & 73.68                             & 82.34                           & 61.25                             & 74.63                           & 62.29          \\
IMN-BERT                                             & 84.06                           & 70.72                             & 77.55                           & 61.73                             & 69.90                           & 60.22          \\
RACL-BERT                                            & 86.38                           & 75.42                             & 81.79                           & 63.40                             & 73.99                           & 66.05          \\
Dual-MRC                                                    & 86.60                           & 75.95                             & 82.51                           & 65.94                             & 75.08                           & 65.08          \\\hline
DiffusionABSA                                        & \textbf{86.17}                           & \textbf{80.64}                    & \textbf{88.37}                  & \textbf{74.90}                    & \textbf{84.62}                  & \textbf{77.26} \\ 
\multicolumn{1}{l}{~~~~~w/o SynTA}                      &85.89                                 &80.11                                   &84.74                                 &70.40                                   &85.62                                 &77.02                \\ \hline 
\hlineB{4}
\end{tabular}}
\caption{Results of AE, AESC on $D_{20b}$ datasets.}
\label{tab:d20b}
\end{table}

\subsection{Performance on Aspects with Various Lengths}
To further assess the effectiveness of \texttt{DiffusionABSA}, we conduct an extensive analysis of its performance across varying aspect lengths. Our evaluation encompasses both our proposed framework and a fine-tuned BERT-large-based sequence labeling model (SeqLab) – a widely acknowledged and robust baseline for ABSA tasks.
Due to space constraints, we present the experimental findings for aspect lengths on datasets 14res and 16res in Table \ref{tab:length}. 
The results reveal the following insights: 1) Our model consistently demonstrates substantial performance enhancements across all aspect lengths. Notably, the average improvement over SeqLab surpasses 14\%; 2) The magnitude of enhancements grows in tandem with the length of the aspect.  
For example, we observe remarkable enhancements exceeding 70\% for datasets 14res and 16res.
All these findings indicate the potency of our framework in facilitating token-level controlled generation, particularly as aspect lengths extend. 

\begin{figure}[!t]
\centering
\includegraphics[width=0.47\textwidth]{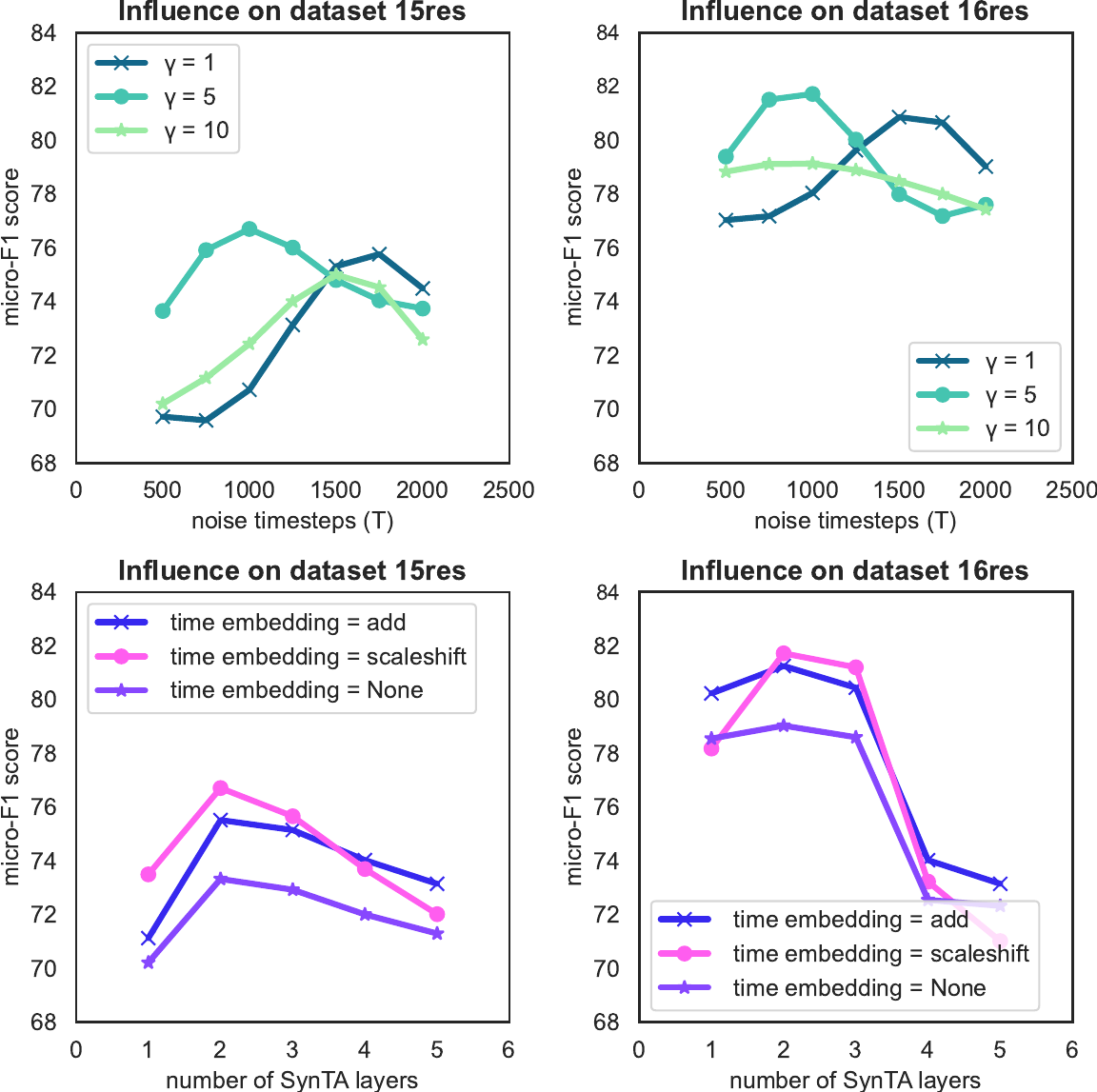} 
\caption{Further analysis of \texttt{DiffusionABSA}.}
\label{fig:ablation}
\end{figure}

\subsection{Further Analysis}
\paragraph{Influence of Hyper-Parameter $\gamma$.} Furthermore, our investigation extends to the examination of \texttt{DiffusionABSA}'s performance in relation to the DDIMs sampling parameter $\gamma$ during the reverse process, as well as the noise timesteps denoted as T within the forward process. Figure \ref{fig:ablation} vividly illustrates the outcomes of this ablation study. As the value of $\gamma$ increases, there is a discernible convergence of the aspects' boundaries towards the ground truth. Particularly compelling is the fact that the most optimal performance is achieved when $\gamma = 5$ and $T = 1000$, specifically within the context of the AESC task. This empirical evidence underscores the effectiveness of our approach in progressively refining aspects' boundaries by leveraging increased sampling $\gamma$, ultimately culminating in a performance zenith characterized by greater alignment with the ground truth.

\paragraph{Influence of Number of SynTA Layers.} 
To assess the efficacy of \texttt{DiffusionABSA}, we conducted a series of experiments to analyze its performance across varying numbers of SynTA layers. Our findings indicate that employing two layers yields superior results compared to a solitary layer, while employing excessively large values may potentially lead to overfitting. 

\paragraph{Influence of Time Embedding.} 
We delve into the impact of time embedding with various operations.
Notably, we observe promising outcomes through the implementation of the scale and shift technique on time embeddings, as illustrated in Figure \ref{fig:ablation}. A significant advantage of this method lies in its capacity to enhance results without substantially augmenting computational overhead.

\section{Conclusions and Future Work}
In this paper, we propose \texttt{DiffusionABSA} equipped with a syntax-aware temporal attention mechanism, which adapts the diffusion model to enhance ABSA by refining the aspect progressively through a dynamic interplay of corruption and denoising process. 
Through comprehensive experimentation across eight benchmark datasets, we empirically validate that \texttt{DiffusionABSA} excels over the compared strong baselines (e.g., ChatGPT), setting a new performance benchmark within the domain. 
Notably, the controlled generation inherent to diffusion models exhibits remarkable efficacy in facilitating the extraction of aspect terms with extended length. 
Specially, for aspects with lengths greater than two, our approach surpasses the SeqLab model with an impressive 70\% enhancement in terms of F1 score.
In the future, we will explore the potential of diffusion models to tackle more complex subtasks such as pair and triplet extraction. 

\section*{Acknowledge}
The authors wish to thank the reviewers for their helpful comments and suggestions. This work was partially funded by National Natural Science Foundation of China (No.62307028), Shanghai Science and Technology Innovation Action Plan (No.23ZR1441800 and No.23YF1426100).

\section*{References}
\bibliographystyle{lrec-coling2024-natbib}
\bibliography{lrec-coling2024-example}

\begin{thebibliography}{45}
\expandafter\ifx\csname natexlab\endcsname\relax\def\natexlab#1{#1}\fi

\bibitem[{Brown et~al.(2020)Brown, Mann, Ryder, Subbiah, Kaplan, Dhariwal, Neelakantan, Shyam, Sastry, Askell et~al.}]{brown2020language}
Tom Brown, Benjamin Mann, Nick Ryder, Melanie Subbiah, Jared~D Kaplan, Prafulla Dhariwal, Arvind Neelakantan, Pranav Shyam, Girish Sastry, Amanda Askell, et~al. 2020.
\newblock Language models are few-shot learners.
\newblock \emph{Advances in neural information processing systems}, 33:1877--1901.

\bibitem[{Chen and Qian(2020)}]{chen2020relation}
Zhuang Chen and Tieyun Qian. 2020.
\newblock \href {https://doi.org/10.18653/v1/2020.acl-main.340} {Relation-aware collaborative learning for unified aspect-based sentiment analysis}.
\newblock In \emph{Proceedings of the 58th Annual Meeting of the Association for Computational Linguistics}, pages 3685--3694, Online. Association for Computational Linguistics.

\bibitem[{Devlin et~al.(2019)Devlin, Chang, Lee, and Toutanova}]{devlin2018bert}
Jacob Devlin, Ming-Wei Chang, Kenton Lee, and Kristina Toutanova. 2019.
\newblock \href {https://doi.org/10.18653/v1/N19-1423} {{BERT}: Pre-training of deep bidirectional transformers for language understanding}.
\newblock In \emph{Proceedings of the 2019 Conference of the North {A}merican Chapter of the Association for Computational Linguistics: Human Language Technologies, Volume 1 (Long and Short Papers)}, pages 4171--4186, Minneapolis, Minnesota. Association for Computational Linguistics.

\bibitem[{Dumoulin et~al.(2018)Dumoulin, Perez, Schucher, Strub, Vries, Courville, and Bengio}]{dumoulin2018feature}
Vincent Dumoulin, Ethan Perez, Nathan Schucher, Florian Strub, Harm~de Vries, Aaron Courville, and Yoshua Bengio. 2018.
\newblock Feature-wise transformations.
\newblock \emph{Distill}, 3(7):e11.

\bibitem[{Fan et~al.(2018)Fan, Feng, and Zhao}]{fan2018multi}
Feifan Fan, Yansong Feng, and Dongyan Zhao. 2018.
\newblock \href {https://doi.org/10.18653/v1/D18-1380} {Multi-grained attention network for aspect-level sentiment classification}.
\newblock In \emph{Proceedings of the 2018 Conference on Empirical Methods in Natural Language Processing}, pages 3433--3442, Brussels, Belgium. Association for Computational Linguistics.

\bibitem[{Fan et~al.(2019)Fan, Wu, Dai, Huang, and Chen}]{fan2019target}
Zhifang Fan, Zhen Wu, Xin-Yu Dai, Shujian Huang, and Jiajun Chen. 2019.
\newblock \href {https://doi.org/10.18653/v1/N19-1259} {Target-oriented opinion words extraction with target-fused neural sequence labeling}.
\newblock In \emph{Proceedings of the 2019 Conference of the North {A}merican Chapter of the Association for Computational Linguistics: Human Language Technologies, Volume 1 (Long and Short Papers)}, pages 2509--2518, Minneapolis, Minnesota. Association for Computational Linguistics.

\bibitem[{Fei et~al.(2022)Fei, Li, Li, Wu, Li, and Ji}]{fei2022inheriting}
Hao Fei, Fei Li, Chenliang Li, Shengqiong Wu, Jingye Li, and Donghong Ji. 2022.
\newblock Inheriting the wisdom of predecessors: A multiplex cascade framework for unified aspect-based sentiment analysis.
\newblock In \emph{Proceedings of the Thirty-First International Joint Conference on Artificial Intelligence, IJCAI}, pages 4096--4103.

\bibitem[{Han et~al.(2023)Han, Peng, Yang, Wang, Liu, and Wan}]{han2023information}
Ridong Han, Tao Peng, Chaohao Yang, Benyou Wang, Lu~Liu, and Xiang Wan. 2023.
\newblock \href {https://arxiv.org/abs/2305.14450} {Is information extraction solved by chatgpt? an analysis of performance, evaluation criteria, robustness and errors}.
\newblock \emph{ArXiv preprint}, abs/2305.14450.

\bibitem[{He et~al.(2022)He, Sun, Wang, Huang, and Qiu}]{he2022diffusionbert}
Zhengfu He, Tianxiang Sun, Kuanning Wang, Xuanjing Huang, and Xipeng Qiu. 2022.
\newblock \href {https://arxiv.org/abs/2211.15029} {Diffusionbert: Improving generative masked language models with diffusion models}.
\newblock \emph{ArXiv preprint}, abs/2211.15029.

\bibitem[{Ho et~al.(2020)Ho, Jain, and Abbeel}]{ho2020denoising}
Jonathan Ho, Ajay Jain, and Pieter Abbeel. 2020.
\newblock \href {https://proceedings.neurips.cc/paper/2020/hash/4c5bcfec8584af0d967f1ab10179ca4b-Abstract.html} {Denoising diffusion probabilistic models}.
\newblock In \emph{Advances in Neural Information Processing Systems 33: Annual Conference on Neural Information Processing Systems 2020, NeurIPS 2020, December 6-12, 2020, virtual}.

\bibitem[{Hu et~al.(2019)Hu, Peng, Huang, Li, and Lv}]{hu2019open}
Minghao Hu, Yuxing Peng, Zhen Huang, Dongsheng Li, and Yiwei Lv. 2019.
\newblock \href {https://doi.org/10.18653/v1/P19-1051} {Open-domain targeted sentiment analysis via span-based extraction and classification}.
\newblock In \emph{Proceedings of the 57th Annual Meeting of the Association for Computational Linguistics}, pages 537--546, Florence, Italy. Association for Computational Linguistics.

\bibitem[{Li et~al.(2020)Li, Chen, Quan, Ling, and Song}]{li2020conditional}
Kun Li, Chengbo Chen, Xiaojun Quan, Qing Ling, and Yan Song. 2020.
\newblock \href {https://doi.org/10.18653/v1/2020.acl-main.631} {Conditional augmentation for aspect term extraction via masked sequence-to-sequence generation}.
\newblock In \emph{Proceedings of the 58th Annual Meeting of the Association for Computational Linguistics}, pages 7056--7066, Online. Association for Computational Linguistics.

\bibitem[{Li et~al.(2022)Li, Thickstun, Gulrajani, Liang, and Hashimoto}]{li2022diffusion}
Xiang Li, John Thickstun, Ishaan Gulrajani, Percy~S Liang, and Tatsunori~B Hashimoto. 2022.
\newblock Diffusion-lm improves controllable text generation.
\newblock \emph{Advances in Neural Information Processing Systems}, 35:4328--4343.

\bibitem[{Li et~al.(2019{\natexlab{a}})Li, Bing, Li, and Lam}]{li2019unified}
Xin Li, Lidong Bing, Piji Li, and Wai Lam. 2019{\natexlab{a}}.
\newblock \href {https://doi.org/10.1609/aaai.v33i01.33016714} {A unified model for opinion target extraction and target sentiment prediction}.
\newblock In \emph{The Thirty-Third {AAAI} Conference on Artificial Intelligence, {AAAI} 2019, The Thirty-First Innovative Applications of Artificial Intelligence Conference, {IAAI} 2019, The Ninth {AAAI} Symposium on Educational Advances in Artificial Intelligence, {EAAI} 2019, Honolulu, Hawaii, USA, January 27 - February 1, 2019}, pages 6714--6721. {AAAI} Press.

\bibitem[{Li et~al.(2019{\natexlab{b}})Li, Bing, Zhang, and Lam}]{li2019exploiting}
Xin Li, Lidong Bing, Wenxuan Zhang, and Wai Lam. 2019{\natexlab{b}}.
\newblock \href {https://doi.org/10.18653/v1/D19-5505} {Exploiting {BERT} for end-to-end aspect-based sentiment analysis}.
\newblock In \emph{Proceedings of the 5th Workshop on Noisy User-generated Text (W-NUT 2019)}, pages 34--41, Hong Kong, China. Association for Computational Linguistics.

\bibitem[{Liu et~al.(2019)Liu, Ott, Goyal, Du, Joshi, Chen, Levy, Lewis, Zettlemoyer, and Stoyanov}]{liu2019roberta}
Yinhan Liu, Myle Ott, Naman Goyal, Jingfei Du, Mandar Joshi, Danqi Chen, Omer Levy, Mike Lewis, Luke Zettlemoyer, and Veselin Stoyanov. 2019.
\newblock \href {https://arxiv.org/abs/1907.11692} {Roberta: A robustly optimized bert pretraining approach}.
\newblock \emph{ArXiv preprint}, abs/1907.11692.

\bibitem[{Ma et~al.(2018)Ma, Li, and Wang}]{ma2018joint}
Dehong Ma, Sujian Li, and Houfeng Wang. 2018.
\newblock \href {https://doi.org/10.18653/v1/D18-1504} {Joint learning for targeted sentiment analysis}.
\newblock In \emph{Proceedings of the 2018 Conference on Empirical Methods in Natural Language Processing}, pages 4737--4742, Brussels, Belgium. Association for Computational Linguistics.

\bibitem[{Ma et~al.(2019)Ma, Li, Wu, Xie, and Wang}]{ma2019exploring}
Dehong Ma, Sujian Li, Fangzhao Wu, Xing Xie, and Houfeng Wang. 2019.
\newblock \href {https://doi.org/10.18653/v1/P19-1344} {Exploring sequence-to-sequence learning in aspect term extraction}.
\newblock In \emph{Proceedings of the 57th Annual Meeting of the Association for Computational Linguistics}, pages 3538--3547, Florence, Italy. Association for Computational Linguistics.

\bibitem[{Mao et~al.(2021)Mao, Shen, Yu, and Cai}]{mao2021joint}
Yue Mao, Yi~Shen, Chao Yu, and Longjun Cai. 2021.
\newblock \href {https://ojs.aaai.org/index.php/AAAI/article/view/17597} {A joint training dual-mrc framework for aspect based sentiment analysis}.
\newblock In \emph{Thirty-Fifth {AAAI} Conference on Artificial Intelligence, {AAAI} 2021, Thirty-Third Conference on Innovative Applications of Artificial Intelligence, {IAAI} 2021, The Eleventh Symposium on Educational Advances in Artificial Intelligence, {EAAI} 2021, Virtual Event, February 2-9, 2021}, pages 13543--13551. {AAAI} Press.

\bibitem[{Mitchell et~al.(2013)Mitchell, Aguilar, Wilson, and Van~Durme}]{mitchell2013open}
Margaret Mitchell, Jacqui Aguilar, Theresa Wilson, and Benjamin Van~Durme. 2013.
\newblock \href {https://aclanthology.org/D13-1171} {Open domain targeted sentiment}.
\newblock In \emph{Proceedings of the 2013 Conference on Empirical Methods in Natural Language Processing}, pages 1643--1654, Seattle, Washington, USA. Association for Computational Linguistics.

\bibitem[{Nachmani and Dovrat(2021)}]{nachmani2021zero}
Eliya Nachmani and Shaked Dovrat. 2021.
\newblock \href {https://arxiv.org/abs/2111.01471} {Zero-shot translation using diffusion models}.
\newblock \emph{ArXiv preprint}, abs/2111.01471.

\bibitem[{Nichol et~al.(2022)Nichol, Dhariwal, Ramesh, Shyam, Mishkin, McGrew, Sutskever, and Chen}]{nichol2022glide}
Alexander~Quinn Nichol, Prafulla Dhariwal, Aditya Ramesh, Pranav Shyam, Pamela Mishkin, Bob McGrew, Ilya Sutskever, and Mark Chen. 2022.
\newblock \href {https://proceedings.mlr.press/v162/nichol22a.html} {{GLIDE:} towards photorealistic image generation and editing with text-guided diffusion models}.
\newblock In \emph{International Conference on Machine Learning, {ICML} 2022, 17-23 July 2022, Baltimore, Maryland, {USA}}, volume 162 of \emph{Proceedings of Machine Learning Research}, pages 16784--16804. {PMLR}.

\bibitem[{OpenAI(2023)}]{openai2023}
OpenAI. 2023.
\newblock Introducing chatgpt.
\newblock \emph{OpenAI Blog.}

\bibitem[{Peng et~al.(2020)Peng, Xu, Bing, Huang, Lu, and Si}]{peng2020knowing}
Haiyun Peng, Lu~Xu, Lidong Bing, Fei Huang, Wei Lu, and Luo Si. 2020.
\newblock \href {https://aaai.org/ojs/index.php/AAAI/article/view/6383} {Knowing what, how and why: {A} near complete solution for aspect-based sentiment analysis}.
\newblock In \emph{The Thirty-Fourth {AAAI} Conference on Artificial Intelligence, {AAAI} 2020, The Thirty-Second Innovative Applications of Artificial Intelligence Conference, {IAAI} 2020, The Tenth {AAAI} Symposium on Educational Advances in Artificial Intelligence, {EAAI} 2020, New York, NY, USA, February 7-12, 2020}, pages 8600--8607. {AAAI} Press.

\bibitem[{Perez et~al.(2018)Perez, Strub, De~Vries, Dumoulin, and Courville}]{perez2018film}
Ethan Perez, Florian Strub, Harm De~Vries, Vincent Dumoulin, and Aaron Courville. 2018.
\newblock Film: Visual reasoning with a general conditioning layer.
\newblock In \emph{Proceedings of the AAAI conference on artificial intelligence}, volume~32.

\bibitem[{Reid et~al.(2022)Reid, Hellendoorn, and Neubig}]{reid2022diffuser}
Machel Reid, Vincent~J Hellendoorn, and Graham Neubig. 2022.
\newblock \href {https://arxiv.org/abs/2210.16886} {Diffuser: Discrete diffusion via edit-based reconstruction}.
\newblock \emph{ArXiv preprint}, abs/2210.16886.

\bibitem[{Schouten and Frasincar(2015)}]{schouten2015survey}
Kim Schouten and Flavius Frasincar. 2015.
\newblock Survey on aspect-level sentiment analysis.
\newblock \emph{IEEE transactions on knowledge and data engineering}, 28(3):813--830.

\bibitem[{Sohl{-}Dickstein et~al.(2015)Sohl{-}Dickstein, Weiss, Maheswaranathan, and Ganguli}]{sohl2015deep}
Jascha Sohl{-}Dickstein, Eric~A. Weiss, Niru Maheswaranathan, and Surya Ganguli. 2015.
\newblock \href {http://proceedings.mlr.press/v37/sohl-dickstein15.html} {Deep unsupervised learning using nonequilibrium thermodynamics}.
\newblock In \emph{Proceedings of the 32nd International Conference on Machine Learning, {ICML} 2015, Lille, France, 6-11 July 2015}, volume~37 of \emph{{JMLR} Workshop and Conference Proceedings}, pages 2256--2265. JMLR.org.

\bibitem[{Song et~al.(2021)Song, Meng, and Ermon}]{song2020denoising}
Jiaming Song, Chenlin Meng, and Stefano Ermon. 2021.
\newblock \href {https://openreview.net/forum?id=St1giarCHLP} {Denoising diffusion implicit models}.
\newblock In \emph{9th International Conference on Learning Representations, {ICLR} 2021, Virtual Event, Austria, May 3-7, 2021}. OpenReview.net.

\bibitem[{Vaswani et~al.(2017)Vaswani, Shazeer, Parmar, Uszkoreit, Jones, Gomez, Kaiser, and Polosukhin}]{DBLP:conf/nips/VaswaniSPUJGKP17}
Ashish Vaswani, Noam Shazeer, Niki Parmar, Jakob Uszkoreit, Llion Jones, Aidan~N. Gomez, Lukasz Kaiser, and Illia Polosukhin. 2017.
\newblock \href {https://proceedings.neurips.cc/paper/2017/hash/3f5ee243547dee91fbd053c1c4a845aa-Abstract.html} {Attention is all you need}.
\newblock In \emph{Advances in Neural Information Processing Systems 30: Annual Conference on Neural Information Processing Systems 2017, December 4-9, 2017, Long Beach, CA, {USA}}, pages 5998--6008.

\bibitem[{Wang et~al.(2017)Wang, Pan, Dahlmeier, and Xiao}]{wang2017coupled}
Wenya Wang, Sinno~Jialin Pan, Daniel Dahlmeier, and Xiaokui Xiao. 2017.
\newblock \href {http://aaai.org/ocs/index.php/AAAI/AAAI17/paper/view/14441} {Coupled multi-layer attentions for co-extraction of aspect and opinion terms}.
\newblock In \emph{Proceedings of the Thirty-First {AAAI} Conference on Artificial Intelligence, February 4-9, 2017, San Francisco, California, {USA}}, pages 3316--3322. {AAAI} Press.

\bibitem[{Wei et~al.(2022)Wei, Wang, Schuurmans, Bosma, Xia, Chi, Le, Zhou et~al.}]{wei2022chain}
Jason Wei, Xuezhi Wang, Dale Schuurmans, Maarten Bosma, Fei Xia, Ed~Chi, Quoc~V Le, Denny Zhou, et~al. 2022.
\newblock Chain-of-thought prompting elicits reasoning in large language models.
\newblock \emph{Advances in Neural Information Processing Systems}, 35:24824--24837.

\bibitem[{Wu et~al.(2020)Wu, Ying, Zhao, Fan, Dai, and Xia}]{wu2020grid}
Zhen Wu, Chengcan Ying, Fei Zhao, Zhifang Fan, Xinyu Dai, and Rui Xia. 2020.
\newblock \href {https://doi.org/10.18653/v1/2020.findings-emnlp.234} {Grid tagging scheme for aspect-oriented fine-grained opinion extraction}.
\newblock In \emph{Findings of the Association for Computational Linguistics: EMNLP 2020}, pages 2576--2585, Online. Association for Computational Linguistics.

\bibitem[{Xu et~al.(2020)Xu, Li, Lu, and Bing}]{xu2020position}
Lu~Xu, Hao Li, Wei Lu, and Lidong Bing. 2020.
\newblock Position-aware tagging for aspect sentiment triplet extraction.
\newblock \emph{arXiv preprint arXiv:2010.02609}.

\bibitem[{Yan et~al.(2021)Yan, Dai, Ji, Qiu, and Zhang}]{yan2021unified}
Hang Yan, Junqi Dai, Tuo Ji, Xipeng Qiu, and Zheng Zhang. 2021.
\newblock \href {https://doi.org/10.18653/v1/2021.acl-long.188} {A unified generative framework for aspect-based sentiment analysis}.
\newblock In \emph{Proceedings of the 59th Annual Meeting of the Association for Computational Linguistics and the 11th International Joint Conference on Natural Language Processing (Volume 1: Long Papers)}, pages 2416--2429, Online. Association for Computational Linguistics.

\bibitem[{Yang and Zhao(2022)}]{yang2022aspect}
Yifei Yang and Hai Zhao. 2022.
\newblock \href {https://aclanthology.org/2022.coling-1.217} {Aspect-based sentiment analysis as machine reading comprehension}.
\newblock In \emph{Proceedings of the 29th International Conference on Computational Linguistics}, pages 2461--2471, Gyeongju, Republic of Korea. International Committee on Computational Linguistics.

\bibitem[{Yu et~al.(2023)Yu, Wu, Li, Bai, and Yang}]{yu2023syngen}
Chengze Yu, Taiqiang Wu, Jiayi Li, Xingyu Bai, and Yujiu Yang. 2023.
\newblock Syngen: A syntactic plug-and-play module for generative aspect-based sentiment analysis.
\newblock In \emph{ICASSP 2023-2023 IEEE International Conference on Acoustics, Speech and Signal Processing (ICASSP)}, pages 1--5. IEEE.

\bibitem[{Yu et~al.(2021)Yu, Ao, Luo, Yang, Sun, Li, and He}]{yu2021making}
Guoxin Yu, Xiang Ao, Ling Luo, Min Yang, Xiaofei Sun, Jiwei Li, and Qing He. 2021.
\newblock Making flexible use of subtasks: A multiplex interaction network for unified aspect-based sentiment analysis.
\newblock In \emph{Findings of the Association for Computational Linguistics: ACL-IJCNLP 2021}, pages 2695--2705.

\bibitem[{Zhang et~al.(2023)Zhang, Zhang, Zhang, and Kweon}]{zhang2023text}
Chenshuang Zhang, Chaoning Zhang, Mengchun Zhang, and In~So Kweon. 2023.
\newblock \href {https://arxiv.org/abs/2303.07909} {Text-to-image diffusion model in generative ai: A survey}.
\newblock \emph{ArXiv preprint}, abs/2303.07909.

\bibitem[{Zhang et~al.(2015)Zhang, Zhang, and Vo}]{zhang2015neural}
Meishan Zhang, Yue Zhang, and Duy-Tin Vo. 2015.
\newblock \href {https://doi.org/10.18653/v1/D15-1073} {Neural networks for open domain targeted sentiment}.
\newblock In \emph{Proceedings of the 2015 Conference on Empirical Methods in Natural Language Processing}, pages 612--621, Lisbon, Portugal. Association for Computational Linguistics.

\bibitem[{Zhang et~al.(2022)Zhang, Li, Deng, Bing, and Lam}]{zhang2022survey}
Wenxuan Zhang, Xin Li, Yang Deng, Lidong Bing, and Wai Lam. 2022.
\newblock A survey on aspect-based sentiment analysis: Tasks, methods, and challenges.
\newblock \emph{IEEE Transactions on Knowledge and Data Engineering}.

\bibitem[{Zhou et~al.(2020{\natexlab{a}})Zhou, Huang, Hu, and He}]{zhou2020sk}
Jie Zhou, Jimmy~Xiangji Huang, Qinmin~Vivian Hu, and Liang He. 2020{\natexlab{a}}.
\newblock Sk-gcn: Modeling syntax and knowledge via graph convolutional network for aspect-level sentiment classification.
\newblock \emph{Knowledge-Based Systems}, 205:106292.

\bibitem[{Zhou et~al.(2020{\natexlab{b}})Zhou, Tian, Wang, Wu, Xiao, and He}]{zhou2020sentix}
Jie Zhou, Junfeng Tian, Rui Wang, Yuanbin Wu, Wenming Xiao, and Liang He. 2020{\natexlab{b}}.
\newblock Sentix: A sentiment-aware pre-trained model for cross-domain sentiment analysis.
\newblock In \emph{Proceedings of the 28th international conference on computational linguistics}, pages 568--579.

\bibitem[{Zhou et~al.(2019)Zhou, Huang, Guo, Han, and Hu}]{zhou2019span}
Yan Zhou, Longtao Huang, Tao Guo, Jizhong Han, and Songlin Hu. 2019.
\newblock \href {https://doi.org/10.24963/ijcai.2019/762} {A span-based joint model for opinion target extraction and target sentiment classification}.
\newblock In \emph{Proceedings of the Twenty-Eighth International Joint Conference on Artificial Intelligence, {IJCAI} 2019, Macao, China, August 10-16, 2019}, pages 5485--5491. ijcai.org.

\bibitem[{Zou et~al.(2023)Zou, Kim, and Kang}]{zou2023diffusion}
Hao Zou, Zae~Myung Kim, and Dongyeop Kang. 2023.
\newblock \href {https://arxiv.org/abs/2305.14671} {Diffusion models in nlp: A survey}.
\newblock \emph{ArXiv preprint}, abs/2305.14671.

\end{thebibliography}

\end{document}